\documentclass[sn-mathphys-num]{sn-jnl}
\usepackage{graphicx}%
\usepackage{multirow}%
\usepackage{amsmath,amssymb,amsfonts}%
\usepackage{amsthm}%
\usepackage{mathrsfs}%
\usepackage[title]{appendix}%
\usepackage{xcolor}%
\usepackage{textcomp}%
\usepackage{manyfoot}%
\usepackage{booktabs}%
\usepackage{algorithm}%
\usepackage{algorithmicx}%
\usepackage{algpseudocode}%
\usepackage{listings}%
\usepackage{amsmath}
\usepackage{algorithm}
\usepackage{algpseudocode}

\usepackage{amsmath}
\usepackage{array}
\usepackage{geometry}
\theoremstyle{thmstyleone}%
\theoremstyle{thmstyletwo}%

\theoremstyle{thmstylethree}%

\begin{document}

\title[Can Deep Learning Trigger Alerts from Mobile-Captured Images?]{Can Deep Learning Trigger Alerts from Mobile-Captured Images?}


\author*[]{\fnm{Pritisha} \sur{Sarkar}}\email{ps.19cs1110@phd.nitdgp.ac.in}

\author[]{\fnm{Duranta Durbaar Vishal} \sur{Saha}}\email{durantadurbaarvishalsaha@gmail.com}

\author[]{\fnm{Mousumi} \sur{Saha}}\email{msaha.cse@nitdgp.ac.in}

\affil*[]{\orgdiv{Computer Science and Engineering}, \orgname{NIT Durgapur}, \orgaddress{\street{M.G.Road}, \city{Durgapur}, \postcode{713209}}}

% \affil[2]{\orgdiv{Department}, \orgname{Organization}, \orgaddress{\street{Street}, \city{City}, \postcode{10587}, \state{State}, \country{Country}}}

% \affil[3]{\orgdiv{Department}, \orgname{Organization}, \orgaddress{\street{Street}, \city{City}, \postcode{610101}, \state{State}, \country{Country}}}


\abstract {Our research presents a comprehensive approach to leveraging mobile camera image data for real-time air quality assessment and recommendation. We develop a regression-based Convolutional Neural Network model and tailor it explicitly for air quality prediction by exploiting the inherent relationship between output parameters. As a result, the Mean Squared Error of 0.0077 and 0.0112 obtained for 2 and 5 pollutants respectively outperforms existing models. Furthermore, we aim to verify the common practice of augmenting the original dataset with a view to introducing more variation in the training phase. It is one of our most significant contributions that our experimental results demonstrate minimal accuracy differences between the original and augmented datasets. Finally, a real-time, user-friendly dashboard is implemented which dynamically displays the Air Quality Index and pollutant values derived from captured mobile camera images. Users' health conditions are considered to recommend whether a location is suitable based on current air quality metrics. Overall, this research contributes to verification of data augmentation techniques, CNN-based regression modelling for air quality prediction, and user-centric air quality monitoring through mobile technology. The proposed system offers practical solutions for individuals to make informed environmental health and well-being decisions.}

\keywords{Image Data Augmentation, Regression-based Convolutional Neural Networks, Air Quality Prediction, Mobile Camera Image Data,
Real-time Dashboard}



\maketitle

\section{Introduction}\label{1_sec}
Addressing air pollution is essential for improving public health and preventing millions of premature deaths worldwide\footnote{\url{http://surl.li/lbnfd}}. The rise of global air pollution due to rapid urbanization and industrialization has become a critical concern, with pollutants like CO, CO$_{2}$, NO$_{2}$, and PM$_{2.5}$ posing severe health risks, including respiratory diseases and reduced lung function. With urbanization projected to reach 70\% by 2050\footnote{\url{https://shorturl.at/mnrLS}}, efficient air quality monitoring is crucial. Current methods, relying on costly stations and sensors, are inadequate. North America and Australia have lower pollution levels than India and Asia, while South America and Africa face higher yet underreported pollution.

Today, numerous devices and platforms are available to monitor current air pollutant levels which may provide route recommendations. However, there is no existing solution that offers real-time pollutant level assessment tailored to an individual's specific health symptoms, identifying which pollutants present in the air are harmful. Our contribution addresses this gap by introducing a novel approach that uses images captured by mobile cameras to determine real-time pollutant levels and assess their impact based on personal health conditions. This innovative method represents a significant advancement in environmental health monitoring.

We propose using mobile camera images as a cost-effective alternative to air pollution sensors to address individual air pollution-related health concerns. Utilizing data from the most polluted cities, we developed a predictive model to recommend whether a current location is suitable for a person’s health. Our previous published research \cite{mine}  is to demonstrate the effectiveness of a Modified CNN model in simultaneously predicting AQI along with pollutant concentrations across three metropolitan cities.
Unlike traditional methods relying on historical sensor data, our current approach addresses challenges like class imbalance and noisy data, resulting in a more robust and accurate model. Sensors are expensive and require high maintenance, so our model's ability to recommend suitability based on a person’s symptoms is highly beneficial. Our model predicts pollutants in real time using image data. With the increasing availability of portable cameras and smartphones, analyzing images for air quality metrics like PM, NO$_{2}$, SO$_{2}$, and CO, along with AQI, offers an efficient and affordable way to monitor air quality. This approach is gaining momentum, with growing literature on computer vision and deep learning for analyzing real time image.

Existing research includes a CNN-based image classifier integrated with a regression model \cite{real}. Zhou et al. \cite{deep16} introduced a probabilistic dynamic causal model for PM$_{2.5}$ dynamics. Deleawe et al. \cite{deep15} explored machine learning for urban CO$_{2}$ level prediction. Another study \cite{image} classified images into three types using CNN, focusing on PM$_{2.5}$ concentration. A novel approach \cite{air} used CNN to predict air pollution from camera images. \cite{last} discussed two methods for image classification: one using KNN and random forest, the other using CNN. Transfer learning, effective in remote sensing, was shown in \cite{air18}. 
So the \textbf{problem statement} is:
%\textbf{Problem Statement:}
{\emph{Can mobile camera image data be feasibly used to recommend health concerns through a single model?}}
Our major \textbf{{contributions}} are:
\vspace{-1.7mm}
\begin{itemize} 
 
\item \textbf{Analysis of Data Augmentation Methods :}
Implemented a robust data augmentation technique to increase input data size effectively and verified the changes in results. Demonstrated minimal accuracy difference between original and augmented data, ensuring reliability.

\item \textbf{Regression-Based CNN Model for Air Quality Prediction :} Designed and implemented a specialized CNN model for predicting air quality parameters. Achieved superior accuracy compared to existing methods by exploiting the inherent relationship between output parameters.

\item \textbf{Real-Time User-Friendly Dashboard - "HealhCamCNN":}
Created a user-friendly dashboard capable of real-time display of Air Quality Index and pollutant values. Utilized captured mobile camera images to provide personalized air quality assessments. Enabled straightforward recommendations regarding location suitability based on current air quality conditions and user health considerations.
\end{itemize}
%\vspace{-1.7mm}

The paper is organised as follows. Section \ref{2_sec} illustrates related research before Section \ref{3_sec} describes an overview of our proposed model. Next, Section  \ref{4_sec} presents the constructed data setup while the methodology is explained in Section \ref{5_sec} . Section \ref{6_sec} then analyses the experimental results and depicts the health recommendation system 'HealthCamCNN', which precedes the conclusion and future work in Section \ref{7_sec}.

\section{Literature Survey}\label{2_sec}
 Air quality forecasting has a rich history, predominantly relying on statistical and shallow machine-learning techniques
   \cite{deep3} such as Regression \cite{deep5} , ARIMA \cite{deep17} , HMM \cite{deep4} , and ANN \cite{deep16}. 
% Zhou et al. \cite{deep16} introduced a probabilistic dynamic causal (PDC) model for PM$_{2.5}$ dynamics. In another study, Deleawe et al.\cite{deep15} explored machine learning for urban CO$_{2}$ level prediction.
Transfer learning works well for remote sensing classification, as demonstrated in \cite{air18}, where an effective strategy involves unsupervised pre-training.
This ability is validated in other research, \cite{air19}, where CNNs excel in operational emergency scenarios for automated building damage assessment. Transfer learning works well in aerial scene classification studies \cite{air20, air21}, with fine-tuned neural networks used for feature extraction and SVM-based remote sensing image classification using linear and RBF kernels. Transfer learning struggles with class-imbalanced datasets, as demonstrated in \cite{air22}, where a pre-trained network performed poorly in detecting invasive blueberry species in wetland aerial images. In her research \cite{improve3}, Athira employed deep learning models, specifically RNN, LSTM, and GRU, to predict future PM$_{10}$ levels using AirNet pollution and weather time series data, with GRU yielding the best results. In their study  \cite{improve4}, Chau proposed Weather Normalized Models (WNM) based on deep learning techniques, including CNN, LSTM, RNN, Bi-RNN, and GRU, to assess air quality variations during the partial COVID-19 shutdown in Quito, Ecuador. Prediction of air quality using multimodal deep learning model combining a Long Short Term Memory (LSTM) and Convolutional Neural Network (CNN) in Seoul \cite{app}. Another research work,\cite{deep}, used two deep neural networks: one-dimensional CNNs and Bi-directional LSTM. This paper.
Salman \cite{improve5} introduced an LSTM-based weather prediction model for the 
Indonesian airport area, incorporating intermediate variable signals within LSTM memory blocks, and demonstrating that the use of the pressure variable, enhanced predictive performance, with the multilayer LSTM model emerging as the most effective. Bekkar \cite{improve6} used a hybrid CNN-LSTM model for PM$_{2.5}$ prediction in Beijing, outperforming LSTM, Bi-LSTM, GRU, and Bi-GRU; their approach integrated spatial data, unlike prior studies. Xie \cite{improve7} introduced a multimodal deep learning model combining CNN and GRU for 6-hour PM$_{2.5}$ predictions in Wuxi using data from the local environmental agency, by leveraging CNN for spatial trends and GRU for long-term dependencies. Kalajdjieski  \cite{improve8} advocated using a custom pre-trained inception model to classify air pollution contamination in Skopje, Macedonia, utilizing multimodal data from sensors and cameras. The research \cite{revise1} enhances PM$_{2.5}$ estimation with a high-resolution AOD algorithm and a regional GWR model, improving daily concentration estimates in Beijing, Tianjin, and Hebei. \cite{revise2}  significantly advances air quality analysis in China by introducing a GTWR model, revealing nuanced spatiotemporal pollutant correlations. 
 Paper \cite{revise3}  advances $PM_{2.5}$ prediction by evaluating Himawari-8's AOD data, highlighting spatial-temporal AOD patterns. \cite{revise4} analyzed water quality management with a high-resolution random forest model for chlorophyll-a prediction, aiding in eutrophication prevention and decision-making processes. The study \cite{revise5} significantly advances dam structure monitoring by employing Ground-based Synthetic Aperture Radar (GB-SAR) data analysis, enhancing regional-scale monitoring with time series displacement data in dynamic environments. Another research study \cite{revise6} enhances our comprehension of soil moisture dynamics in eastern China from 1961 to 2011, using GLDAS data. It uncovers seasonal variations, drying trends, and the influence of temperature and precipitation on various soil layers vital for future drought anticipation and water resource management. In paper \cite{revise7}, the author represents dam deformation monitoring and prediction using the Lévy flight bat optimization algorithm (LBA) and hybrid-kernel extreme learning machine (KELM). Survey summarization of some existing research work is represented in Table \ref{survey}.

\begin{table}[]
\tiny
\caption{Survey summarization of existing research work}\label{survey}
\begin{tabular}{|l|l|l|l|l|l|}
\hline
\textbf{Ref.} & \textbf{Model}    & \textbf{Contribution}   &           \textbf{Result}     & \textbf{City, Country}    \\ \hline

\cite{new1}       & 3D-CNN GRU                                                                                                 & \begin{tabular}[c]{@{}l@{}}Attention layers for feature \\ extraction\end{tabular}                                            & \begin{tabular}[c]{@{}l@{}}Comparison with 3D-CNN,\\  CNN \& CNN-LSTM\end{tabular}                     & Shanghai, China                                                                 \\ \hline

\cite{real}      & \begin{tabular}[c]{@{}l@{}}CNN-RC, \\    \\ RN \& VN Scheme\end{tabular}                                   & \begin{tabular}[c]{@{}l@{}}Estimating PM$_{2.5}$, PM$_{10}$ \\ \& AQI with a single model\end{tabular}                                  & \begin{tabular}[c]{@{}l@{}}R\_Squared = 76\%-84\%,   \\ Daytime vs. night-time\\ analysis\end{tabular} & \begin{tabular}[c]{@{}l@{}}Kaohsiung City, \\ Taiwan\end{tabular}   \\ \hline

\cite{deep}     & \begin{tabular}[c]{@{}l@{}}1D-CNN \& Bi-\\ directional LSTM\end{tabular}                                   & \begin{tabular}[c]{@{}l@{}}Forecasting PM$_{2.5}$ using \\ multivariate time series data\end{tabular}                              & \begin{tabular}[c]{@{}l@{}}DAQFF has least RMSE \& \\ MAE.\end{tabular}                                & Beijing, China          \\ \hline

\cite{image}               & \begin{tabular}[c]{@{}l@{}}Imagenet-matconvnet-\\ very deep model, random\\  forest classifier\end{tabular} & \begin{tabular}[c]{@{}l@{}}First method that applies \\ CNN for image-based PM$_{2.5}$\\ estimation with 591   images\end{tabular} & \begin{tabular}[c]{@{}l@{}}Random Forest-63\%, \\ CNN - 68\%\end{tabular}                              & Beijing, China                                                                                  \\ \hline

\cite{hardini1}               & CNN, SEM-PLS analysis                                                                                      & AIR-Protection Platform                                                                                                       & \begin{tabular}[c]{@{}l@{}}Reliability tests, 257\\ questionnaires\end{tabular}                        & Indonesia                                                        \\ \hline

\cite{nature1}               & Spatial Autocorrelation                                                                                    & \begin{tabular}[c]{@{}l@{}}AQI prediction using \\ lockdown data\end{tabular}                                                 & \begin{tabular}[c]{@{}l@{}}47\% \& 67\% reduction \\ in prediction error\end{tabular}                  & \begin{tabular}[c]{@{}l@{}}Shanghai \& \\ Wuhan, China\end{tabular}  \\ \hline

\cite{improve6}               & \begin{tabular}[c]{@{}l@{}}Hybrid CNN-LSTM\\  multivariate\end{tabular}                                    & \begin{tabular}[c]{@{}l@{}}Predict hourly forecast of \\ PM$_{2.5}$ using meteorological\\ data (no images used)\end{tabular}      & \begin{tabular}[c]{@{}l@{}}Outperforms existing \\ combinations of CNN \\ \& LSTM\end{tabular}         & Beijing, China        \\ \hline

\cite{last}                & \begin{tabular}[c]{@{}l@{}}CNN \& Gabor\\  Transform\end{tabular}                                          & \begin{tabular}[c]{@{}l@{}}Prediction using \\ smartphone images\end{tabular}                                                 & 59.38\%                                                                                                & Tehran, Iran       \\ \hline
\end{tabular}
\end{table}

The research gap in prior papers lies in their limited focus on specific regions, challenges with class-imbalanced datasets, and suboptimal integration of multimodal data for air quality prediction. Our research addresses these gaps by presenting a novel approach using a modified CNN named 'HealthCamCNN' model. We provide a comprehensive analysis of air quality across diverse polluted cities in India, utilizing both pollution and image data. This broad and integrated approach distinguishes our research, making it a significant advancement in the field of air quality forecasting.

\section{System Overview}\label{3_sec}

\begin{figure*}[h!]
\begin{center}
\includegraphics[width=12cm]{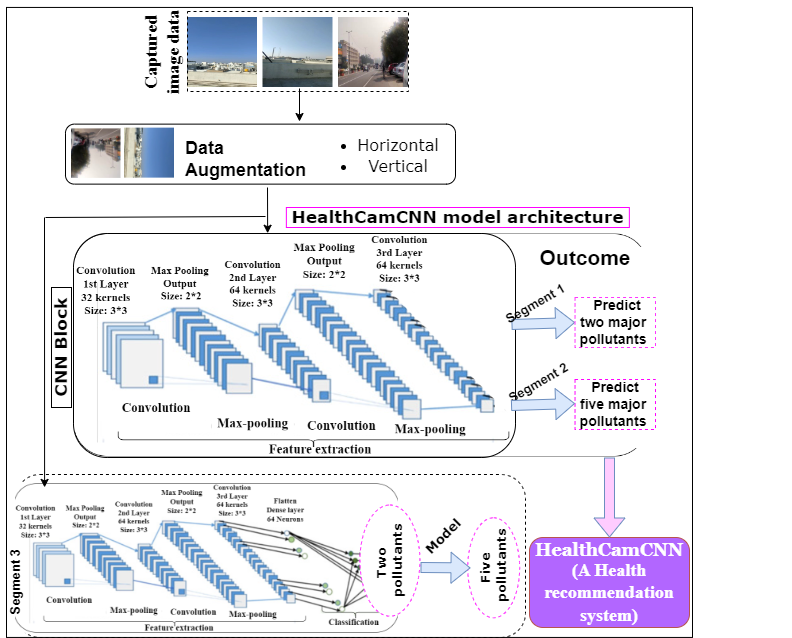}
\caption{Workflow of our proposed approach}
\label{diwali}
\end{center}
\end{figure*}

Our research paper aims to leverage historical datasets from three major regions of Bengaluru, Delhi and Tamil Nadu to predict air quality. Our approach involves a systematic series of steps represented in Figure \ref{diwali}:

1. \textbf{Data collection:} We collected historical datasets for the aforementioned cities, which include various air quality parameters such as PM$_{2.5}$, PM$_{10}$, SO$_{2}$, O$_{3}$, NO$_{2}$, and CO, along with corresponding timestamps.

2. \textbf{Image data preprocessing:} The RGB images obtained were preprocessed using statistical mechanisms and converted to numpy arrays ready to be fed as inputs to a neural network. Data augmentation was also implemented using horizontal and vertical transformations so as to analyse their effect on the obtained results.

3. \textbf{Modified CNN model:} We employed a modified CNN model, 'HealthCamCNN' to extract key features from the 3D arrays. This CNN model is customized with specific parameter values and internal mechanisms tailored for our air quality prediction task.

4. \textbf{Accuracy assessment:} We evaluated the performance  models separately to gain key insights. 

In the first part of our method, we used a CNN model to predict two pollutants $PM_{2.5}$ and $PM_{10}$. Then we applied it to predict the remaining five pollutants from real-time polluted images separately. In our modified CNN model(in Figure \ref{diwali} purple boundaries part),'HealthCamCNN' model we first predict two pollutants from images, and then predict five additional pollutants. This modified version is implemented in a dashboard system, allowing users to easily assess their exposure to pollutants based on their symptoms and receive recommendations on whether the current environmental conditions are suitable for them.

This evaluation includes measuring the accuracy of each model in predicting air quality levels. We performed validation exercises to ensure the robustness of our models and predictions. Validation helps confirm the reliability of the results obtained from both the image and binary data models.

Through this comprehensive approach, we gained insights into the effectiveness of using image data for predicting air quality levels. By comparing the accuracy of our models, we determined the potential of image data as a valuable source for air quality predictions, in addition to traditional numerical datasets. This system overview highlights our research's key steps and methodologies to achieve accurate air quality predictions for multiple cities.

\section{Data  Description} \label{4_sec}
We collected a dataset comprising 5455 daytime images from Bengaluru, Delhi, and Tamil Nadu, paired with corresponding $PM_{2.5}$ and $PM_{10}$ values from 2023. Captured under diverse weather conditions, the images included significant portions of the sky and objects at various depths, thereby facilitating air quality assessment. The data were categorized into 6 classes, namely Good, Moderate, Unhealthy for Sensitive Groups, Unhealthy, Very Unhealthy and Severe based on $PM_{2.5}$ levels, together with the AQI, $PM_{2.5}$, $SO_{2}$, $O_{3}$, and $NO_{2}$ values. RGB images were converted to numpy arrays before being scaled to floating-point values by dividing by 255.

\subsection{Data Augmentation}
Data augmentation is performed to increase the diversity and size of training datasets, which helps improve the robustness and generalization ability of machine learning models by exposing them to a wider range of variations and scenarios present in real-world data.

\begin{table}[h!]
\centering
\begin{tabular}{|c|l|}
\hline
\textbf{Notation} & \textbf{Description} \\ \hline
$I$ & Original image \\ \hline
$x$ & Length of the image \\ \hline
$y$ & Width of the image \\ \hline
$m$ & Midpoint along the length of the image, $m = \frac{x}{2}$ \\ \hline
$I_1$ & Left half of the image after vertical splitting \\ \hline
$I_2$ & Right half of the image after vertical splitting \\ \hline
$I_{\text{mirrored}}$ & Resultant image after vertical mirroring \\ \hline
\end{tabular}
\caption{Notation Table}{\label{not1}}
\end{table}

\begin{algorithm}
\caption{ \textbf{Vertical Splitting and Mirroring of Original Image}}\label{alg1}
\begin{algorithmic}[1]
\State \textbf{Begin}
\State Identify the dimensions of the image: $x$ (length) and $y$ (width).
\State Calculate the midpoint: $m = \frac{x}{2}$.
\State \textbf{\textit{Vertical Splitting :}} Divide the image vertically at the midpoint to obtain two new images:
\State \hspace{1em} First image $I_1$: dimensions $(0, m) \times y$
\State \hspace{1em} Second image $I_2$: dimensions $(m, x) \times y$
\State \textbf{\textit{Mirroring:}} Flip the image $I$ with respect to the vertical line along the midpoint such that the rightmost point of the original image becomes the leftmost point in the resultant image $I_{\text{mirrored}}$:
\[
I_{\text{mirrored}}(i, j) = I(x - i, j) \quad \text{for } 0 \leq i < x \text{ and } 0 \leq j < y
\]
\State \textbf{End}
\end{algorithmic}
\end{algorithm}

\begin{enumerate}
   
\item \textbf{Vertical Splitting and Reflecting:} 
As per the Algorithm 1 (with Notation Table \ref{not1}), each image is split vertically into two sub-images. Both sub-images are then reflected vertically, resulting in a total of four images derived from each original image. This process effectively quadruples the size of our training set, thereby augmenting it.

\item \textbf{Maintaining Output Values:} The augmented images retain the same output values as the original image. This ensures that the training labels remain consistent with the new images.

\item \textbf{Shuffling the Dataset :} The dataset is shuffled to introduce randomness. Shuffling helps reduce inherent biases present in the sequence of entries, leading to a more robust and generalizable model.

\item \textbf{Exploring Horizontal Reflection:} Horizontal reflection is sometimes avoided \cite{image} as it would alter the position of the sky in the images, and preserving the natural orientation of the sky is considered crucial for maintaining the contextual integrity of the images. We test this hypothesis by implementing horizontal reflection and analyse the variation in the loss values.
\end{enumerate}

This augmentation strategy enhances the variability of training data, thereby potentially improving the robustness and generalization of our model.

\section{Methodology } \label{5_sec}
Convolution, a generalization of dot product for higher dimensions, has proved useful for extracting key features from images.  The kernel scans the image horizontally from left to right and the output of a convolution operation is always a single real number. Therefore, when applied to a greyscale image, the output will always be a 2D array. Because our inputs are 3-channel RGB images, each kernel used will produce a separate 2D convolution output, all of which are concatenated to obtain the Feature Map for the first layer.
For every non-overlapping sub-sample of a predefined size of the feature map, maxpooling retains only the maximum entry. 
Finally, since we aim to predict the real numbered values of the parameters as opposed to a range, we flatten the final feature map such that the number of neurons= (length* width* number of channels) of the map. A multilayer perceptron consisting of one hidden layer is then implemented such that the number of neurons in the output layer correspond to number of output parameters. An activation function is used to prevent diminishing gradients. The discrepancy is then calculated using a suitable error function so as to obtain more accurate predictions in the following iteration.  

\subsection{Steps of our proposed method : }   
As per the Algorithm 2 (with Notation Table \ref{not2}), the following steps are followed in our methodology;
\begin{enumerate}
    \item  \textbf{Data Compilation:} We have sorted the dataset such that images of a given AQI class are grouped. Created a structured dataset by pairing each image array with its corresponding label.
    \item  \textbf{Feature-Label Separation:} Extracted features (image pixel representations) and labels from the dataset: - Features $(X)$: 3D arrays representing the images. - Labels $(Y)$: Corresponding classification labels.
    \item  \textbf{Neural Network Configuration:} Utilized the Keras library for building a sequential neural network.
    \item  \textbf{Layer Importation:} Imported essential layers such as Conv2D, MaxPooling2D, Dense, and Flatten.
    \item   \textbf{Convolutional Layers:} Implemented 3 Convolutional layers, each comprising 1 Conv2D layer followed by 1 MaxPooling2D layer, with 32 filters for the first layer \& 64 for the later ones, each of size 3*3.  We used LeakyReLU activation function to alleviate the problem of terminating ReLU \cite{vision}.
   
    \item   \textbf{Model branching:} As shown in Figure \ref{diwali} in segment 1 predict two pollutants and in segment 2 predict 5 pollutants.
    
    \item   \textbf{Flattening:} Flattened the output from the convolutional layers,in segment 3 predict two pollutants from images and then predict another 5 pollutants from these two pollutant.

\end{enumerate}

\begin{table}[ht]
    \centering
    \caption{Notation Table for Model Architecture}
    \begin{tabular}{|c|p{10cm}|}
        \hline
        \textbf{Symbol} & \textbf{Meaning} \\
        \hline
        \( I \) & Input images of size \( I_{\text{height}} \times I_{\text{width}} \) \\
        \hline
        \( F^{(i)} \) & Filter set \( i \) with specific sizes for convolution \\
        \hline
        \( Z^{(i)} \) & Output of convolution operation for stage \( i \) \\
        \hline
        \( A^{(i)} \) & Output of LeakyReLU activation for stage \( i \) \\
        \hline
        \( P^{(i)} \) & Output of maxpooling operation for stage \( i \) \\
        \hline
        \( F \) & Flattened feature vector for input to MLPs \\
        \hline
        \( Y^{(1)} \) & Predictions for pollutants $PM_{2.5}$ and $PM_{10}$ \\
        \hline
        \( Y^{(2)} \) & Predictions for remaining parameters \\
        \hline
        \( \text{Conv}(X, F) \) & Convolution operation of input \( X \) with filter \( F \) \\
        \hline
        \( \text{LeakyReLU}(X) \) & LeakyReLU activation function applied to \( X \) \\
        \hline
        \( \text{Maxpool}(X) \) & Maxpooling operation applied to \( X \) \\
        \hline
        \( \text{Flatten}(X) \) & Flatten operation to convert \( X \) into a vector \\
        \hline
        \( \text{MLP}_1, \text{MLP}_2 \) & Multilayer Perceptrons for segment 1 and segment 2 \\
        \hline
    \end{tabular}\label{not2}
\end{table}

\begin{algorithm}
\caption{Model Architecture for Pollutants based on $PM_{2.5}$ and $PM_{10}$}\label{alg2}
\begin{algorithmic}[1]
    \State \textbf{Input:} Images $I$ of size $I_{\text{height}} \times I_{\text{width}}$
    \State \textbf{Initialization:} Define filter sets $F^{(1)}, F^{(2)}, F^{(3)}$ with increasing sizes
    
    \State \textbf{Convolution and Activation:}
    \State $Z^{(1)} = \text{Conv}(I, F^{(1)})$ \Comment{Convolution operation}
    \State $A^{(1)} = \text{LeakyReLU}(Z^{(1)})$ \Comment{LeakyReLU activation}
    
    \State \textbf{Maxpooling:}
    \State $P^{(1)} = \text{Maxpool}(A^{(1)})$ \Comment{Maxpooling operation}
    
    \State \textbf{Iterative Feature Extraction:}
    \For{$i = 2$ to $3$} \Comment{Iterate twice more}
        \State $Z^{(i)} = \text{Conv}(P^{(i-1)}, F^{(i)})$ \Comment{Convolution with $F^{(i)}$}
        \State $A^{(i)} = \text{LeakyReLU}(Z^{(i)})$ \Comment{Activation}
        \State $P^{(i)} = \text{Maxpool}(A^{(i)})$ \Comment{Maxpooling}
    \EndFor

    \State \textbf{Model Branching:}
    \State \textbf{Segment 1:} Predict pollutants $PM_{2.5}$ and $PM_{10}$ \\
    \hspace{\algorithmicindent} $Y^{(1)} = \text{MLP}_1(F)$
    
    \State \textbf{Segment 2:} Predict remaining parameters \\
    \hspace{\algorithmicindent} $Y^{(2)} = \text{MLP}_2(F)$

       \State \textbf{Flattening for MLP Input:}
    \State $F = \text{Flatten}(P^{(3)})$ \Comment{Flatten into vector}
    
     \State \textbf{Segment 3: HealthCamCNN model architechture} Predict two pollutants based on image data, then estimate remaining pollutants using the predicted values of two pollutants as input
    
    \State \textbf{Output:} Prediction of pollutants values
\end{algorithmic}
\end{algorithm}

The image processing begins with convolution operations on the input images, each followed by a LeakyReLU activation layer to introduce non-linearity. Subsequently, maxpooling is applied to retain the most significant features extracted by each convolutional kernel. This convolution and maxpooling process is iterated three times, progressively increasing the number of filters to capture more complex patterns in the images. After convolution, the resulting kernels are flattened into vectors to facilitate input into subsequent multilayer perceptrons (MLPs). The model architecture bifurcates into two segments: the first segment predicts two pollutants, $PM_{2.5}$ and $PM_{10}$, while the second segment predicts the remaining five parameters. In the third segment in our modified CNN method we predict 2 pollutant from images firstly the predict 5 pollutant from that 2 pollutant. This methodology describes the detailed steps, including data pre-processing and CNN model training, in a more concise and organized manner.

\section{Results and Discussion}\label{6_sec}
We implement a Convolutional Neural Network equipped with Max Pooling between two Convolution layers and Regression layers at the end to predict the output parameters. Given all images are RGB of size 224*224, we use 32 filters in the first layer \& 64 for the following two layers, all of size 3*3, as shown in figure \ref{diwali}. A pool size of 2*2 is used for max-pooling without any stride or padding. Adam optimizer is utilized to compile the model \& both Mean Absolute Error (MAE) \& Mean Squared Error (MSE) were used to assess the performance of the model.
Each of the models were run for 50 epochs. The model was implemented using Tensorflow \& Python and trained on a Windows 10 machine with Intel core i5 CPU and 16gb RAM.
Finally, we deployed our model to a real-time dashboard that can display all the output parameters upon uploading a new image. The dashboard thus makes our work user-friendly and practical as it facilitates air quality prediction solely from captured images without necessitating the use of sensors.

We explored two separate neural network models exploiting their inherent architectures to obtain better results for the respective parameters. The first model included two segments with $PM_{2.5}$ and $PM_{10}$ being predicted in the first stage preceded by the convolution block. The predicted values obtained thereby are then used as inputs to a multilayer perceptron which predicts the values of the remaining 5 parameters. The results obtained thus outperforms the second model which branches the 2 groups of parameters after the convolution block. The resultant model not only has less overhead but also performs significantly better than the one simultaneously predicting all the 7 parameters.

{The testing error values for the two-stage model have been noted in Table \ref{stages} and the branching model in Table \ref{two},\ref{newtab} respectively. It is obvious that the latter outperforms the former, thereby establishing the branched CNN model as the superior architecture.
}

% \begin{figure*}
% % \begin{center}
% \includegraphics[width=15cm]{norm.drawio.png}
% \caption{Region-wise CNN accuracy, A$_{1}$, B$_{1}$, C$_{1}$ denotes Bengaluru, A$_{1}$ and B$_{1}$ denote epochs vs. the Mean Absolute Error graph of the entire model and of each parameter, respectively, before normalization, while C$_{1}$ illustrates the performance post-normalisation. Similarly, A$_{2}$, B$_{2}$, C$_{2}$ for Delhi and A$_{3}$, B$_{3}$, C$_{3}$ for Tamil Nadu.}
% \label{graphimage}
% % \end{center}
% \end{figure*}

\begin{table}[]
\begin{tabular}{|l|l|l|l|}
\hline
                                   & \textbf{Bengalore} & \textbf{Delhi} & \textbf{Tamil Nadu} \\ \hline
\textbf{Mean Absolute Error(MAE)} & 0.1812             & 0.2787         & 0.2521               \\ \hline
\textbf{Mean Squared Error(MSE)}  & 0.0570             & 0.1135         & 0.0981               \\ \hline

\end{tabular}
\caption{MSE and MAE values for predicting two pollutants from images and five pollutants from two, across Bangalore, Delhi, and Tamil Nadu}\label{stages}
\end{table}

\begin{table}[]
\begin{tabular}{|l|l|l|l|}
\hline
                                   & \textbf{Bengalore} & \textbf{Delhi} & \textbf{Tamil Nadu} \\ \hline
\textbf{Mean Absolute Error(MAE)} &  0.0951            &     0.0559    & 0.0851               \\ \hline
\textbf{Mean Squared Error(MSE)}  & 0.0189             & 0.0077        & 0.0208               \\ \hline

\end{tabular}
\caption{MSE and MAE values for predicting two pollutants from images across Bangalore, Delhi, and Tamil Nadu}\label{two}
\end{table}

\begin{table}[]
\begin{tabular}{|l|l|l|l|}
\hline
                                   & \textbf{Bengalore} & \textbf{Delhi} & \textbf{Tamil Nadu} \\ \hline
\textbf{Mean Absolute Error(MAE)} &   0.1289           &     0.0671     & 0.0946              \\ \hline
\textbf{Mean Squared Error(MSE)}  &  0.0295            &      0.0112    & 0.0226               \\ \hline

\end{tabular}
\caption{MSE and MAE values for predicting five pollutants from images across Bangalore, Delhi, and Tamil Nadu}\label{newtab}
\end{table}

As a result, all the output parameters range between [0,1] during training and are then unnormalized using the inverse transform to compare the predicted values on the test set.

\subsection{Data augmentation results}

\begin{figure*}
 % \begin{center}
\includegraphics[width=15cm]{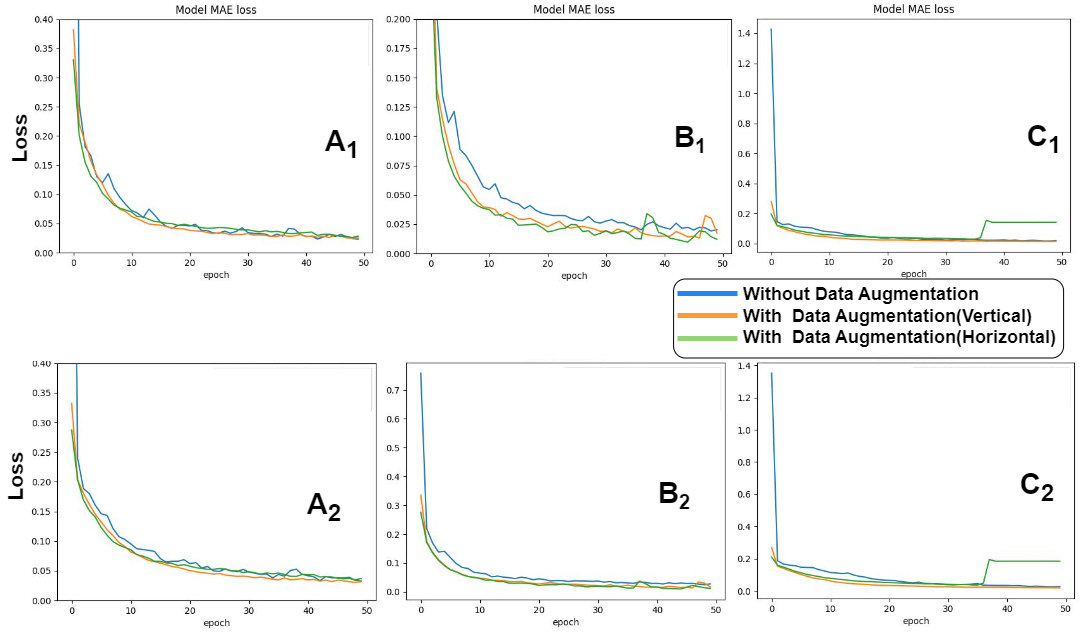}
\caption{Data augmentation visualization : A Tamil Nadu B Delhi; $A_{1}$ and $A_{2}$  - followed the segment 1 method, $B_{1}$ and $B_{2}$ - followed the segment 2 method, $C_{1}$ and $C_{2}$ - followed the segment 3 method,}
\label{aug}
 % \end{center}
\end{figure*}

The Mean Absolute Error obtained with our model with and without data augmentation is included in Figure \ref{aug}. Here, Horizontal Data Augmentation includes a horizontal reflection of the input images, thereby allowing the sky to feature in the bottom, whereas Vertical Data Augmentation does not. First of all, we note that it barely makes any difference whether or not we include the horizontally reflected images in our training set. Therefore the assumption made in \cite{image} to skip this method of data augmentation is invalid. Secondly, we make the remarkable observation that even though the original images without data augmentation always start with a much higher error compared to the augmented ones, the eventual value upon reaching a plateau is almost indistinguishable. Therefore we can henceforth let go of the common practice in CNN models of augmenting the datasets safe in the knowledge that it makes little difference to the end result. We establish this as one of the most significant results obtained in our work.     

% \begin{figure*}
% % \begin{center}
% \includegraphics[width=10cm]{NEW123.drawio.png}
% \caption{Air quality prediction results for PM$_{2.5}$, PM$_{10}$, O$_{3}$, CO, SO$_{2}$, NO$_{2}$) along with AQI across Bengaluru, Delhi and Tamil Nadu. The images in each subfigure depict our model's best and worst output with each parameter's predicted and actual value.}
% \label{cityimage}
% % \end{center}
% \end{figure*}

% Our proposed model was also tested on a dataset comprising 146 night-time images captured by mobile phones. But the obtained mean absolute error is 0.2097 of that captured night images, almost twice as high as that obtained with daytime images of the original dataset. Therefore our future work will include expanding the current model for night-time images to obtain satisfactory accuracy.  

% \begin{figure*}
% % \begin{center}
% \includegraphics[width=11cm]{KJK.drawio (1).png}
% \caption{Some snapshot of our real-time AQI predictor dashboard}
% \label{snap}
% % \end{center}
% \end{figure*}

\subsection{HealthCamCNN: A Real-Time Health Recommendation System}
\begin{figure*}
 \begin{center}
\includegraphics[width=15cm]{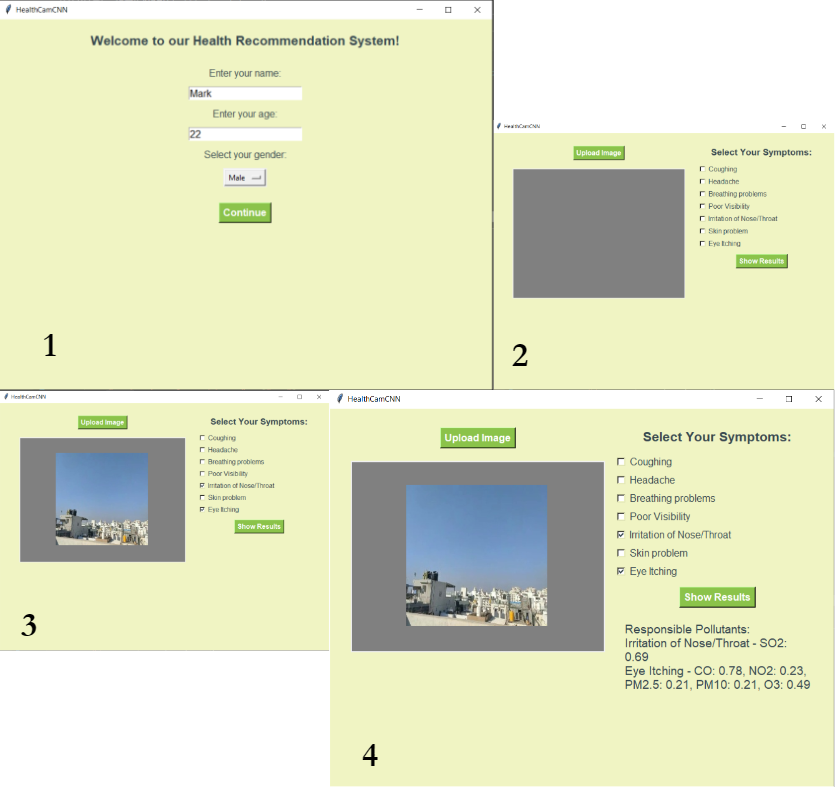}
\caption{Some snapshot of real-time health-related symptoms used to predict AQI in our HealthCamCNN dashboard}
\label{dash}
 \end{center}
\end{figure*}

We have developed an innovative dashboard (the snapshots are represented in Figure \ref{dash} for better understanding) that utilizes image recognition technology to assess current air quality levels for various pollutants, providing users with valuable information conveniently. Unlike traditional methods involving expensive and maintenance-intensive physical devices, our application offers a cost-effective solution. Users can simply upload images of their surroundings, and the dashboard will analyze and present real-time data on air pollutant levels, accompanied by an AQI reading. This recommendation tool proves invaluable for individuals planning to venture outdoors, assisting them in identifying suitable locations based on air quality conditions.

%%%%%%%%%%%%  new ----
The dashboard Figure \ref{dash} enables users to upload images for immediate assessment of air quality at specific locations. It not only provides real-time air quality data derived from uploaded images but also recommends whether these locations are suitable based on the user's health conditions.

\section{Conclusion} \label{7_sec}
This research has successfully developed and validated a robust framework for utilizing mobile camera image data in real-time air quality assessment and recommendation systems. Key contributions include introducing an effective data augmentation method that expands input data size without compromising prediction accuracy. Experimental results demonstrate minimal differences in accuracy between original and augmented datasets, affirming the reliability of the augmentation approach. Moreover, a regression-based CNN model was specifically engineered to accurately predict pollutant levels from image data. The CNN model outperformed existing methods, showcasing its efficacy in providing precise environmental assessments. A pivotal aspect of this study is the implementation of a user-friendly, real-time dashboard that dynamically presents the Air Quality Index and pollutant values derived from mobile camera images. Tailored recommendations on location suitability based on current air quality metrics and user health profiles enhance decision-making capabilities. Looking forward, further advancements can be explored in several areas. Incorporate user feedback mechanisms to continually refine and improve recommendation accuracy based on real-time user experiences. These future directions aim to advance the capabilities of mobile-based air quality monitoring systems, ultimately contributing to improved environmental health management and decision support for individuals and communities.

\backmatter

\section*{Declarations}
All authors have read, understood, and have complied as applicable with the statement on "Ethical responsibilities of Authors" as found in the Instructions for Authors

\begin{itemize}
 % \item Conflict of interest - {Pritisha Sarkar, Duranta Durbaar Vishal Saha and Mousumi Saha declare that they have no conflict of interest.}
\item Ethics approval - NA
\item Consent to participate - Informed consent was obtained from all individual participants included in the study.
\item Funding - No funding was obtained for this study
\item Consent for publication -  All participants/Authors have consented to the submission of the case report to the journal.
\item Availability of data - NA
\item Code availability - NA
 % \item Authors' contributions  -- {Pritisha Sarkar - Conceptualization of this study, Methodology, Software, Original draft preparation; Duranta Durbaar Vishal Saha - Software, Data acquisition and Methodology and Mousumi Saha - Conceptualization of this study, Methodology, Writing an original draft. }
\end{itemize}

\bibliography{sn-bibliography}% common bib file

\end{document}